\documentclass[12p,a4paper,oneside]{amsart}

\usepackage{amsmath}
\usepackage{amsthm}
\usepackage{graphicx}

\usepackage{apacite}
\usepackage{natbib}

\usepackage{mathpazo}
\usepackage{domitian}
\usepackage[T1]{fontenc}

\usepackage{setspace}
\begin{document}

\title[Adaptive Sampling Policies Imply Biased Beliefs]{Adaptive Sampling Policies Imply Biased Beliefs: \\A Generalization of the Hot Stove Effect}

\author{Jerker Denrell  \\ University of Warwick \\ jdenrell@gmail.com}


\maketitle

\begin{abstract}
The Hot Stove Effect is a negativity bias resulting from the adaptive character of learning. The mechanism is that learning algorithms that pursue alternatives with positive estimated values, but avoid alternatives with negative estimated values, will correct errors of overestimation but fail to correct errors of underestimation. Here, we generalize the theory behind the Hot Stove Effect to settings in which negative estimates do not necessarily lead to avoidance but to a smaller sample size (i.e., a learner selects fewer of alternative B if B is believed to be inferior but does not entirely avoid B). We formally demonstrate that the negativity bias remains in this set-up. We also show there is a negativity bias for Bayesian learners in the sense that most such learners underestimate the expected value of an alternative.

\end{abstract}

\section{Introduction}
Learning from experience does not necessarily generate unbiased beliefs, partly due to psychological biases but also due to biases in the information learners sample and get exposed to. One important bias in sampling is the so-called "hot stove effect" \citep{denmarch} which refers to the asymmetry in error correction generated by adaptive learning processes. 
The key idea is that the tendency to avoid alternatives with unfavorable past outcomes generates a biased set of experiences. Alternatives that are underestimated - believed to be worse than what they are – are unlikely to be tried and sampled again, which implies that errors of underestimation are unlikely to be corrected. Alternatives that are overestimated - believed to be better than what they are - are likely to be tried and sampled again which implies that errors of overestimation are likely to be corrected. This asymmetry in error correction generates a biased set of experiences that, in turn, can give rise to biased judgments \citep{denrell2005}, including in-group bias and apparent risk-averse behavior \citep{denrell2005,denrell2007}. 

There is good experimental support for the hot-stove effect at the individual level and researchers in psychology have relied on the hot-stove effect to explain regularities in risk taking in experimental studies \citep{erevroth} and why people underestimate the trustworthiness of others \citep{fetch}. Researchers in finance \citep{dittmar} have used field data to demonstrate that the hot stove effect can explain the risk-taking behavior of executives. The hot stove effect also has important implications for information aggregation and online reviews: If consumers avoid products with poor reviews and consumers review products they buy, negative reviews will be more persistent than positive reviews, generating biased averages \citep{lemens2018}.

Past theoretical work on the hot stove effect has assumed that negative experiences may lead to avoidance, that is, the alternative has not been tried at all \citep{denrell2005,denrell2007}. Clearly, if no more information is available, a negative impression will persist. In many settings, however, a negative belief or impression may not lead to avoidance of the alternative, but rather to a smaller sample size. An animal with a more favorable impression of the energy content of a plant of type A than of a plant of type B may search for plants of type A. During this search for plants of type A, some plants of type B may be incidentally found. The result is that the animal samples more plants of type A (because the search is focused on such plants) than of type B, but the animal does not avoid plants of type B but simply samples fewer of them.
Similarly, a firm may prefer to hire graduates from university A, but may hire some, although fewer, graduates from university B if there are not enough graduates from A that accept its offers. 

In this paper, I generalize the theory behind the hot-stove effect and show that it holds even if a negative impression only leads to a reduction in the sample size, not necessarily to avoidance. 
Specifically, I show that the final belief will be biased for a broad class of learning algorithms in which the sample size is a function of the past belief. If the sample size is higher and the past belief was more positive, there is a negativity bias: The final belief will be lower than the expected value of the random variable the learner is learning about.
This result also applies to taking averages: the average of a sample will be biased if the total sample size is a function of the average based on an initial subset. The bias will be eliminated in the long run as the number of samples increases, but in the short run it can be significant. 

I also examine whether the bias remains for a Bayesian learner. I show that there is no bias on average for a Bayesian learner: the average belief is equal to the expected value of the random variable the learner is learning about. However, I also show that most Bayesian learners will underestimate the variable they are learning about if the sample size is an increasing function of the initial belief. 

These results imply that a large class of sensible and adaptive learning processes can be expected to generate biased beliefs, even if decision-makers process the available information in a seemingly unbiased way (i.e., taking averages). Indeed, even rational Bayesian learners will tend to underestimate the expected value of an alternative (i.e., most will do so) if the total sample size is higher when the initially observed payoffs are high. 
Adaptive sampling processes are common and often necessary to reduce search costs. It is not sensible, for example, to continue to sample an alternative a fixed number of times if initial trials reveal that this alternative has a much lower payoff than other available alternatives. The results in this paper show that even unbiased processing of information generated by such sampling policies can generate seemingly biased beliefs. Adaptive sampling policies thus offer an alternative explanation of biases in beliefs, such as a tendency to underestimate the extent to which others are trustworthy.

\section{Illustration}
To illustrate the basic ideas, we consider a simple two-period set-up. 
In period one, a learner samples an alternative $k$ times and observes the payoffs generated. That is, the learner observes $k$ payoffs, $x_{1,1},x_{1,2},...,x_{1,k}$, each independently drawn from the payoff distribution $f(x)$, where $f(x)$ is assumed to be a normal distribution with mean zero and variance $\sigma^2$. 
Based on the observed payoff, the learner computes the average observed payoff after the first period: $\bar{x}_{1}  =(1/k) \sum_{j=1}^{k}x_{1,j}$.

In the second period, the learner takes an additional sample and observes $m$ payoffs, $x_{2,1},x_{2,2},...,x_{2,m}$, each independently drawn from the payoff distribution $f(x)$. We assume that the size of this sample, $m$, is a function of $\bar{x}_{1}$. For example, the learner may take a larger sample if the observed average first-period payoff is high ($\bar{x}_{1}$ is high), than if the observed average first-period payoff is low ($\bar{x}_{1}$ is low), because the alternative is believed to be more rewarding when the observed average first-period payoff is high compared to when it is low. 
To illustrate the impact of such an adaptive sample size policy, suppose that the sample size in period two is equal to $m = h (high)$ whenever $\bar{x}_{1} > c$ and equal to $m = l (low)$ whenever $\bar{x_{1}} \leq c$. 

After the second period, the learner computes the average of all the payoffs observed in periods one and two: $\bar{x}_{2} = (1/(k+m))[\sum_{j=1}^{k}x_{1,j}+\sum_{j=1}^{m}x_{2,j}]$. 
We are interested to see whether this average is unbiased or not. 

The answer is that this average will be biased.
To illustrate this, suppose that the learner samples two payoffs in the first period ($k=2$), samples ten more if the first-period average is positive ($h = 10$) but only samples one more if the first-period average is negative ($l = 1$). 
The average of all observed payoffs after the second period will then be negative.
When $\sigma$, the standard deviation of the payoff equals $1$, then $E[\bar{x}_{2}]= -0.141$ and the proportion of negative averages is 0.587. When $\sigma = 5$, $E[\bar{x}_{2}]= -0.705$ and the proportion of negative averages is 0.583. 
More generally, 
\begin{equation}
E[\bar{x}_{2}]= -\frac{\sigma\sqrt{k}(h-l)}{\sqrt{2\pi}(k+l)(k+h)}e^{-c^2k/2\sigma^2},
\end{equation}
(see the appendix for a derivation of this equation). 
This equation shows that the bias depends on how the sample size varies with $\bar{x}_{1}$. If the learner takes a larger sample when $\bar{x}_{1}$ is high compared to when $\bar{x}_{1}$ is low ($h>l$), there is a negative bias: $E[\bar{x}_{2}] < 0$. If the learner instead takes a larger sample when $\bar{x}_{1}$ is low compared to when $\bar{x}_{1}$ is high ($h<l$), there is a positive bias: $E[\bar{x}_{2}] > 0.$
Equation (1) also implies that the bias is greater when the payoffs are more variable, that is, when $\sigma^{2}$ is greater. 
In summary, an average based on an adaptive sampling policy (adaptive in the sense that the sample size is a function of the initial average) will be biased, and the bias depends on the type of sampling policy (increasing or decreasing in the initial average). 
Because learners often regulate sample sizes in response to feedback from initial samples, this implies that learning processes generally lead to biased beliefs, at least if the beliefs are based on the average payoffs observed. 
Note that this occurs even if the decision maker does not process the information in a biased way. The decision maker takes the average of the observed payoffs. This averaging process, which is unbiased when the sample size is fixed, becomes biased when the total sample size depends on the initially observed average.

\section{Intuition}

To understand the reason for the bias, note that the sum of all observations after the second period is the sum of two components: the sum of all observations after the first period ($\sum_{j=1}^{k}x_{1,j}$) and the sum of all observations in the second period ($\sum_{j=1}^{m}x_{2,j}.$). The sample size taken in the second period ($m$) will affect the relative weight of these two components. The reason for the negative bias, when the learner takes a larger sample when $\bar{x}_{1}$ is high compared to when $\bar{x}_{1}$ is low ($h>l$), is that the first component will be weighted relatively more when $\bar{x}_{1}$ is low, which implies that the sample size taken in period two is low ($m = l$).

To illustrate this, suppose that the average after two observations in the first period ($k = 2$) is $\bar{x}_{1} = 1$. The sum of the first two periods was equal to $k\bar{x}_{1} = 2$. 
Suppose $c = 0$, implying that the learner will take a large sample ($m = h$), whenever $\bar{x}_{1} > 0$. Suppose $h = 10$, that is, the sample size is ten in period two if the first period average is positive. The average payoff observed in both periods is then
\begin{equation*}
\bar{x}_{2} = \frac{2+\sum_{j=1}^{10}x_{2,j}}{2+10} = \frac{2}{2+10} +\frac{\sum_{j=1}^{10}x_{i,j}}{2+10}.
\end{equation*}
The expected value of an observation in the second period is zero, that is, $E[x_{2,j} = 0]$. It follows that 
\begin{equation*}
E[\bar{x}_{2}  | \bar{x}_{2} =1] = \frac{2}{2+10} = \frac{1}{6},
\end{equation*}
which is close to zero. A positive first-period average payoff tends to regress to the mean of the distribution, which is zero. The reason is that a large sample is taken in the second period, which implies that the first-period average will not be weighted much.

Suppose, in contrast, that the average after two observations in the first period ($k = 2$) was $\bar{x}_{1} = -1$. The sum of the first two periods was then equal to $k\bar{x}_{1} = -2$. 
When $c = 0$, the learner will take a small sample ($m = l$) in the second period because $\bar{x}_{1} = -1 < c$. Suppose $l = 1$: only one sample is taken if the belief in the first period was negative. The average payoff observed in both periods is then
\begin{equation*}
\bar{x}_{2} = \frac{-2+x_{2,1}}{2+1} = \frac{-2}{2+1} +\frac{x_{2,1}}{2+1}.
\end{equation*}
Again, the expected value of an observation in the second period is zero, i.e. $E[x_{2,1} = 0]$. It follows that 
\begin{equation*}
E[\bar{x}_{2}  | \bar{x}_{2} = -1] = \frac{-2}{3},
\end{equation*}
which is further away from zero than $E[\bar{x}_{2}  | \bar{x}_{2} =1] = \frac{1}{6}$ is. A negative first-period average payoff tends to regress less to the mean of the distribution than a positive first-period average payoff. The reason is that a small sample is taken in the second period if the average in the first period is negative. The negative first-period average will be more weighted in the final average. 

As this example shows, adaptive sampling implies that negative first-period averages are more 'persistent': they are given a larger weight than positive first-period averages.
Because negative and positive first-period averages of the same magnitude are equally likely, when the payoff distribution is normal with mean zero (this distribution is symmetric around zero), the greater persistence of negative first-period averages explains the overall negative bias.

The impact of variance can also be understood in a similar way.  If the payoff distribution is more variable, the first period averages will tend to differ more from zero, both in the positive and negative directions. Positive first-period averages will tend to regress more to the mean (zero) than negative averages. When the negative first-period averages are more extreme because the variance is greater, the result is a stronger bias. 

Of course, the bias will be eliminated in the long run, as the number of samples increases since the average of $n$ samples of a random variable $X$ will converge to its expected value. In the short run, however, the bias can be significant.

\section{General Theorem about Biased Averages}

The bias generated by an adaptive sampling size policy does not only hold for the normal distribution and the specific binary sampling policy considered above (high above zero, low below), but it holds for any distribution and a large class of adaptive sampling policies.  

\medskip

\noindent \textbf{Theorem 1:} In period one, a learner observes $k$ payoffs, $x_{1,1},x_{1,2},...,x_{1,k}$, each independently drawn from the payoff distribution $f(x)$, with expected value $E[x] = u$. In period two, the learner samples $n(\bar{x}_{1})$ payoffs, each independently drawn from the payoff distribution $f(x)$. Here, $n(\bar{x}_{1}) \geq 1$, and $n(\bar{x}_{1})$ is a function of the average payoff of the first period: $\bar{x}_{1}  =(1/k) \sum_{j=1}^{k}x_{1,j}$. Let $\bar{x}_{2}$ be the average payoff observed during periods one and two. Then i) $E[\bar{x}_{2}] < u$ whenever $n(\bar{x}_{1})$ is a strictly increasing function of $\bar{x}_{1}$, and ii) $E[\bar{x}_{2}] > u$ whenever $n(\bar{x}_{1})$ is a strictly decreasing function of $\bar{x}_{1}$. 
 \smallskip

\noindent \textbf{Proof: }See the Appendix. \smallskip

 \smallskip
For unimodal symmetric distributions, centered on the expected value $E[x] = u$, it can also be shown that a majority of the averages, after the second period, are below $u$ if higher averages in the first period lead to larger sample sizes in the second period: 

\medskip
\noindent \textbf{Theorem 2:} In period one, a learner observes $k$ payoffs, $x_{1,1},x_{1,2},...,x_{1,k}$, each independently drawn from the payoff distribution $f(x)$ which is symmetric around its expected value $E[x] = u$ (i.e., $f(x-u) = f(x+u)$. In period two, the learner samples $n(\bar{x}_{1})$ payoffs, each independently drawn from the payoff distribution $f(x)$. Here, $n(\bar{x}_{1}) \geq 1$, and $n(\bar{x}_{1})$ is a function of the average payoff of the first period: $\bar{x}_{1}  =(1/k) \sum_{j=1}^{k}x_{1,j}$. Let $\bar{x}_{2}$ be the average payoff observed during periods one and two. Then i) $P[\bar{x}_{2} > u] < 0.5$ whenever $n(\bar{x}_{1})$ is a strictly increasing function of $\bar{x}_{1}$, and ii) $P[\bar{x}_{2} > u] > 0.5$ whenever $n(\bar{x}_{1})$ is a strictly decreasing function of $\bar{x}_{1}$. 
 \smallskip

\noindent \textbf{Proof: }See the Appendix. \smallskip

\section{Alternative Learning Models}

So far we have assumed that the learner computes the average of the observed payoffs, but a similar bias holds for several other learning models. Suppose, for example, that the learner gives more weight to the most recently observed payoff. 
This results in a similar bias:

\smallskip

\noindent \textbf{Theorem 3:} In period one, a learner observes $k$ payoffs, $x_{1,1},x_{1,2},...,x_{1,k}$, each independently drawn from the payoff distribution $f(x)$, with expected value $E[x] = u$. The learner updates his or her belief $z$ after each observed payoff, giving more weight to the most recently observed payoff. Specifically, if the prior belief was $z_t$, the new belief is $z_{t+1} = (1-b)z_{t}+bx_{t}$, where $b$ is the weight of the most recently observed payoff. The initial belief is assumed to be unbiased: $z_0 = u$. 
In period two, the learner samples $n(z_{1,k})$ payoffs, each independently drawn from the payoff distribution $f(x)$. Here $n(z_{1,k}) \geq 1$, and $n(z_{1,k})$ is a function of belief after the $k$ observed payoffs in the first period, $z_{1,k}$. Let $z_{2}$ be the belief after periods one and two. Then i) $E[z_{2}] < u$ whenever $n(z_{1,k})$ is a strictly increasing function of $\bar{x}_{1}$, and ii) $E[z_{2}] > u$ whenever $n(z_{1,k})$ is a strictly decreasing function of $\bar{x}_{1}$. 
 \smallskip

\noindent \textbf{Proof: }See the Appendix. \smallskip

\section{Bayesian updating}

One might suspect that the bias occurs because the learner ignores the sample size when updating. A rational learner should, after all, update less when the sample size is small and update more if the sample size is large, which would seem to work against the above bias. This leads to the question of whether the bias persists if the learner is a Bayesian updater, who takes the sample size into account when updating, and whose belief is the conditional expectation given the observed data, i.e. $b_2 = E[u | X]$? The answer is that there is no bias on average in the following sense: The expected value of the belief after the two periods, $E[b_{2}]$, is equal to the expected value of the prior. For example, if Bayesian learners learn about the mean of a random variable, $u_i$, and the means are drawn from a prior distribution, $f(u_i)$, with expected value $E[u_i] = m$, then $E[b_{2}] = m$. Averaging over different learners, who draw different values of $u_i$, thus generates no bias. However, the distribution of beliefs may still be ``biased'' (or skewed) in the following sense: a majority of Bayesian updaters will have a belief after two periods below $m$. 

To explain this in more detail, consider a learner who observes independent draws from a random variable with distribution $f_{i}(X)$ with expected value $u_{i}$. The learner knows that $u_{i}$ is drawn from a prior distribution with expected value $m$ ($E[u_i] = m$). Let $b_1$ be the belief after having observed $k$ independent draws $x_{1,1},x_{1,2},...,x_{1,k}$ from $f_{i}(X)$ in the first period,  $b_1= E[u_i|x_{1,1},x_{1,2},...,x_{1,k}]$. Let $b_2$ be the belief after all observations in periods one and two: $b_2= E[u_i|x_{1,1},...,x_{2,n(b_1)}]$. 

It is important to distinguish between two ways that the learner's belief can be 'biased'. Consider first a given value of $u_i = k$. We may ask if the expected belief is unbiased in the sense that $E[b_n|u_i = k] = k$, where $b_n$ is the belief after a fixed sample of $n$ observations (fixed in the sense that the number of observations is independent of the value of the observations). The average is unbiased in this sense, but it is well known that Bayesian updating is not. Suppose, for example, that $u_i$ is drawn from a normal distribution with mean zero and variance one. The learner observes $x_i = u_i+\varepsilon_i$ where $\varepsilon_i$ is drawn from a normal distribution with mean zero and variance one. For this setup, it is well known that the expected value of the posterior after one observation is $x_i/2$ (e.g., DeGroot, 1970). Suppose now $u_i = 5$. The expected value of the belief after one observation, given that  $u_i = 5$, equals $E[x_1/2|u_i = 5] = 2.5$ which is lower than 5. Bayesian updating can be biased in the sense that the expected belief, given a set of observations and $u_i = k$, is not equal to $k$. The reason is that the belief gradually increases from the prior value (zero) towards the expected value (five). 

Bayesian updating is unbiased, however, if we average over all learners, in the following sense. Let $b_i$ denote the belief of a Bayesian learner, i.e., $b_i = E[u_i|X]$, where $X$ is the observed data. Then $E[b_i] = m$ where $m = E[u_i]$ is the expected value of the prior. For example, suppose that $u_i$ is drawn from a normal distribution with mean zero and variance one and the learner observes $x_i = u_i+\varepsilon_i$, where $\varepsilon_i$ is drawn from a normal distribution with mean zero and variance one. We then have $E[b_i] = E[x_i/2] = E[(u_i+\varepsilon_i)/2] = 0$. Thus, $E[b_i] = E[u_i] = 0$.

We now show that Bayesian updating remains unbiased in this latter sense, even if the sample size in period two depends on the belief after period one. 
This is true for any distribution:
\medskip

\noindent \textbf{Theorem 4:} 
In period one the learner observes $k$ independent draws $x_{1,1},x_{1,2},...,x_{1,k}$, from distribution $f_{i}(X)$ with expected value $u_i$. The learner knows that $u_i$ is drawn from a prior distribution, $f(u_i)$, with expected value $E[u_i] = m$.
In period two, the learner observes $n(b_1)$ independent draws from distribution $f_{i}(X)$: $x_{2,1},x_{2,2},...,x_{2,n(b_1)}$. 
The sample size in the second period, $n(b_1)$, is a function of the belief after the first period, $b_1 = E[u_i|x_{1,1},...,x_{1,k}]$. Let $b_2$ be the belief after all observations, in both periods one and two: $b_2= E[u_i|x_{1,1},...,x_{2,n(b_1)}]$.
Then $E[b_2] = m$. \smallskip

\noindent \textbf{Proof of Theorem 4:} 
The proof is simple. We condition on the belief after the first period: $E[b_2|b_1]$. Because conditional expectations are martingales (e.g., \cite{williams1991}, p. 96), we have $E[b_2|b_1] = b_1$. That is, there is no change, in expectation, from the belief after period one to the belief after the information in the second period. This is true since if information that would lead to such a change could be anticipated in period one, it should already have been incorporated into the belief, of a rational agent in period one. From $E[b_2|b_1] = b_1$ it follows that $E[b_2] = E_{b_1}[E[b_2|b_1]] = E[b_1]$.
By the tower property of martingales (e.g., \cite{williams1991}, p.88) we have $E[b_1] = E[E[u_i|x_{1,1},x_{1,2},...,x_{1,k}]]= E[u_i] = m$. Thus, $E[b_2] = m$. 

\medskip
While there is no bias when averaging the beliefs of all learners (i.e. $E[b_2] = m$), most Bayesian learners may have beliefs below $m$ if positive beliefs lead to larger sample sizes. 
To illustrate this, suppose that $f_{i}(X)$ is a normal distribution with mean $u_{i}$ and variance $\sigma_{e}^{2}$. Moreover, suppose that $u_{i}$ is drawn from a normal distribution with mean $m$ and variance $\sigma_{u}^{2}$. The learner observes $k$ independent draws $x_{1,1},x_{1,2},...,x_{1,k}$ from $f_{i}(X)$ in the first period. In period two, the learner observes $n(b_1)$ independent draws from distribution $f_{i}(X)$: $x_{2,1},x_{2,2},...,x_{2,n(b_1)}$. 
\medskip

\noindent \textbf{Theorem 5:} 
If the payoff and the prior distributions are both normal, then i) whenever $n(b_1)$ is a strictly increasing function of $b_1$, most learners will, after the second period, underestimate the random variable they are learning about: $Pr(b_2 < m) > 0.5$.  ii) Whenever $n(b_1)$ is a strictly decreasing function of $b_1$, most learners will, after the second period, overestimate the random variable they are learning about: $Pr(b_2 > m) > 0.5$.  \smallskip

\noindent \textbf{Proof: }See the Appendix. \smallskip

Thus, even if Bayesian learners are learning about a symmetric distribution and they have a prior that is symmetric around zero ($m = 0$), most learners will have a negative belief after sampling. This is due to the adaptive sampling policy. The intuition is similar to why the learning policy that takes the average is biased: negative initial beliefs are more persistent because the learner will then not take many additional samples, and the initial few negative observations will be weighted heavily. Note that such a bias would not occur if the learner followed a fixed sampling policy and decided, at the outset, how many samples to take. If the sample size was fixed, the belief would be equally likely to be positive or negative, when learning about a normally distributed payoff with a mean taken from normally distributed prior with mean zero. Note also that when sampling is adaptive, all Bayesian learners know that only 50\% of the means are negative. Still, most Bayesian learners believe that the mean they observe is negative. 

\begin{figure}
\noindent \begin{centering}

\includegraphics[trim=0cm 9cm 0cm 6cm, clip=true, scale=0.65]{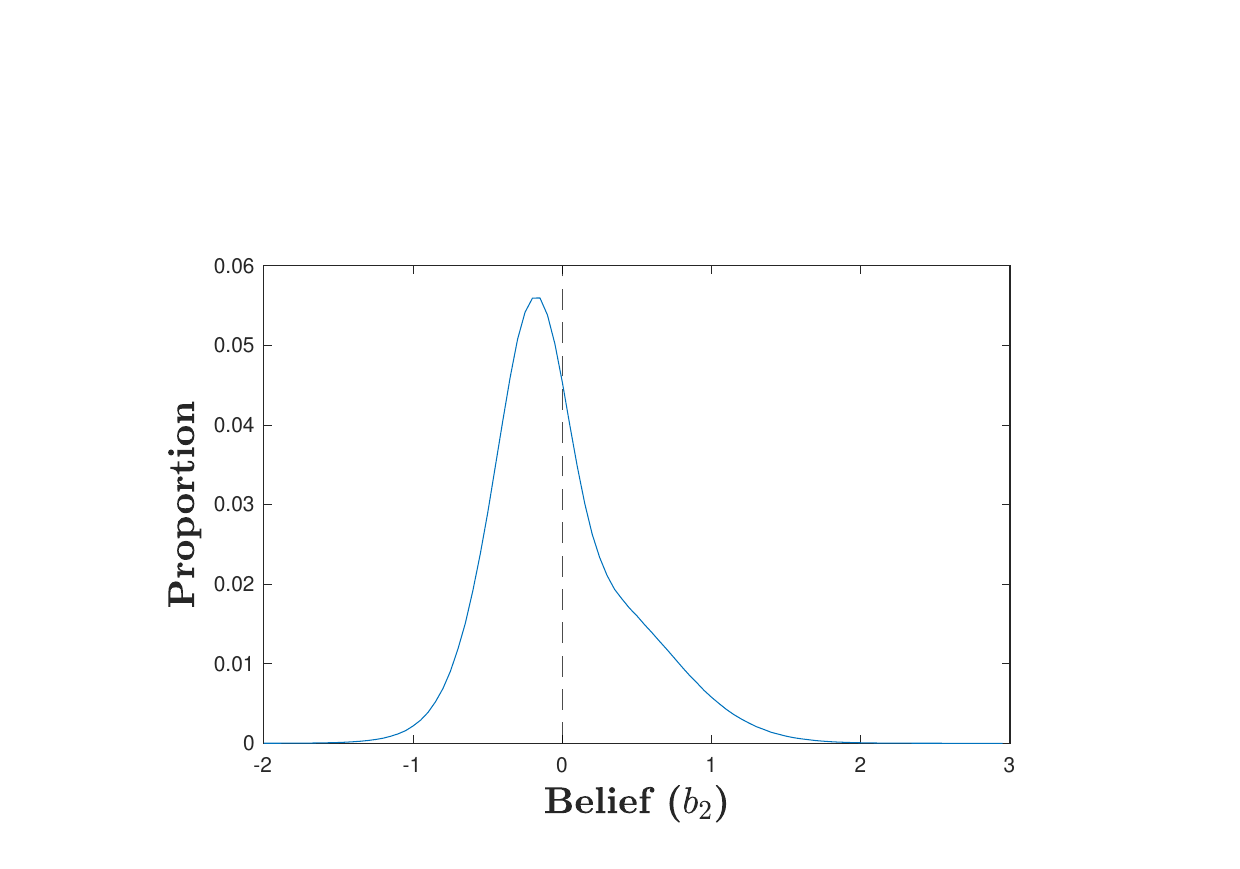}
\caption{The distribution of beliefs of a Bayesian learner after the second period when $\sigma^{2}_{e} = 5$.}

\label{skew}

\par\end{centering}
\end{figure}

To illustrate this, suppose that payoffs are normally distributed with mean $m$ and variance $\sigma^{2}_{e}$ where $m$ is drawn from a normal prior with mean $u = 0$ and variance $\sigma^{2}_{p} = 1$. Consider a learner who samples two payoffs in the first period, samples ten more if the belief after the first period is positive, but only samples one more if the belief after the first period is negative. When $\sigma^{2}_{e} = 1$, the proportion of beliefs after the second period that are negative, $P(b_2 < 0)$ where $b_2 =  E[u_i|x_{1,1},...,x_{2,n(b_1)}]$, is 0.534. When $\sigma^{2}_{e} = 5$, the proportion of beliefs after the second period that are negative is 0.577 (based on 10 million simulations). Figure \ref{skew} shows the distribution of beliefs after the second period for the case where $\sigma^{2}_{e} = 5$.


That a majority of Bayesian learners end up with a negative belief may seem paradoxical since on average there is no bias: $E[b_2] = 0$ when $m = 0$, i.e., the average belief, averaging over all Bayesian learners, is equal to the mean of the prior. 
The paradox is resolved by noting that the learners who believe that the mean is negative are less confident in this estimate than the learners who believe that the mean is positive and therefore took a larger sample in the second period. This results in a skewed distribution of beliefs after the second period. Most beliefs are below $m = 0$ but those below $m = 0$ are less likely to be extreme beliefs than beliefs above $m = 0$ since beliefs below $m = 0$ are based on smaller sample sizes than those above $m = 0$. 
This can be seen in Figure \ref{skew}.

Because Bayesian updating with normal distributions relies on the observed average, it may also seem puzzling that Bayesian updating is unbiased in the sense that $E[b_2] = m$ (Theorem 3) while averaging is biased in the sense that $E[\bar{x} | u_i = k] < k$ for all $k$ (Theorem 1). 
Consider again the case where $u_{i}$ is drawn from a normal distribution with mean $m$ and variance $\sigma_{u}^{2}$. 
In period one, the learner observes $k$ independent draws $x_{1,1},x_{1,2},...,x_{1,k}$, $x_{1,j} = u_i+\varepsilon_j$ where $\varepsilon_j$ are independently drawn from a normal distribution with mean zero and variance $\sigma_{e}^{2}$. In period two, the learner observes $n(b_1)$ independent draws from the same distribution: $x_{2,1},x_{2,2},...,x_{2,n(b_1)}$, where $n(b_1)$ is a function of $b_1 = E[u_i|x_{1,1},...,x_{1,k}]$. 
The Bayesian estimate of $u_i$ given the observations in both periods equals (DeGroot, 1970): $E[u_i|X]=(\bar{x}\sigma_{u}^{2}+(\sigma_{e}^{2}/n_2)m)/(\sigma_{u}^{2}+(\sigma_{e}^{2}/n_2))$, where $n_2 = k+n(b_1)$, the total number of observations in both periods, and $\bar{x}$ is the average of all observations in both periods. 
As $\sigma_{u}^{2} \to \infty$ $E[u_i|X] \to \bar{x}$, that is, converges to the average. How can then the Bayesian estimate be unbiased while the average is biased? The puzzle is resolved by noting that when $\sigma_{u}^{2}$ grows large compared to $\sigma_{e}^{2}$, the signal-to-noise ratio ($\sigma_{u}^{2}/ \sigma_{e}^{2}$) becomes large. When $\sigma_{u}^{2} \to \infty$ the signal-to-noise ratio goes to infinity. The noise term is then vanishingly small compared to the values of $u_i$. Because the bias in the average depends on the possibility that the first-period observations can far below (or above) $u_i$, i.e., depend on noise, there is only a vanishingly small bias in the average when $\sigma_{u}^{2} \to \infty$, resolving the seeming paradox. 

\section{Implications for Understanding Learning}

The bias resulting from adaptive sampling policies implies that even seemingly unbiased learning algorithms, such as averaging or Bayesian updating, can result in biased beliefs, at least in the sense that most learners underestimate (or overestimate) an alternative. This offers an alternative explanation of some judgment biases; an explanation that does not require a psychological bias that assumes biases in information processing. For example, suppose that it is observed that a firm tends to underestimate the productivity of graduates from universities. Such a negativity bias. can be explained by an adaptive sampling policy (the firm tends to hire fewer people from universities it has had a worse experience with) and does not require motivated reasoning or a cognitive bias.

\raggedbottom

\pagebreak

\section*{Appendix}

\noindent \textbf{Derivation of equation 1:} 
\begin{align*}
E[b_2]=P(\bar{x_{1}} >c)E[b_2|\bar{x_{1}}>c]+ P(\bar{x_{1}} <c)E[b_2|\bar{x_{1}}<c].
\end{align*}
We have
\begin{align*}
E[b_2|\bar{x_{1}}>c] = E[\frac{S_2}{k+h}|\bar{x_{1}}>c]+E[\frac{\bar{x_{1}}k}{k+h}|\bar{x_{1}}>c].
\end{align*}
where $S_2$ is the sum of observations in period two. 
Due to independence, $E[\frac{S_2}{k+h}|\bar{x_{1}}>c] = \frac{1}{k+h}E[S_2] = 0$. 
Using the expression for the expectation of a truncated normal distribution (see, e.g., Greene 2000, p. 952) we have 
\begin{align*}
E[\bar{x_{1}}|\bar{x_{1}}>c] = \frac{\sigma_e}{\sqrt{k}}\frac{\phi(\alpha)}{1-\Phi(\alpha)},
\end{align*}
where $\alpha = c/(\sigma_e/\sqrt{k}) = -c\sqrt{k}/\sigma_e$. Moreover, $P(\bar{x_{1}} >c) = 1-\Phi(\alpha)$. Hence,
\begin{align*}
P(\bar{x_{1}} >c)E[b_2|\bar{x_{1}}>c] =  (1-\Phi(\alpha))\frac{k}{k+h}(\frac{\sigma_e}{\sqrt{k}}\frac{\phi(\alpha)}{1-\Phi(\alpha)}) \\
= \frac{\sigma_e\sqrt{k}}{k+h}\phi(\alpha).
\end{align*}
Similarly, 
\begin{align*}
P(\bar{x_{1}} <c)E[b_2|\bar{x_{1}}<c] = \Phi(\alpha)\frac{k}{k+l}(-\frac{\sigma_e}{\sqrt{k}}\frac{\phi(\alpha)}{\Phi(\alpha)}) \\
= -\frac{\sigma_e\sqrt{k}}{k+l}\phi(\alpha).
\end{align*}
Overall we get
\begin{align*}
E[b_2]=\frac{\sigma_e\sqrt{k}}{(k+l)(k+h)}[( (k+l)\phi(\alpha)-(k+h)\phi(\alpha)]\\
=-\frac{\sigma_e\sqrt{k}}{(k+l)(k+h)}[(h-l)\phi(\alpha)].
\end{align*}
Now, $\phi(\alpha) = \frac{1}{\sqrt{2\pi}}e^{-\alpha^2/2}$. Hence,
\begin{align*}
E[b_2]=-\frac{\sigma_e\sqrt{k}(h-l)}{\sqrt{2\pi}(k+l)(k+h)}e^{-c^2k/2\sigma_e^2}.
\end{align*}

\smallskip

\noindent \textbf{Proof of Theorem 1:} 
The average of all payoffs observed in periods one and two is, given the average observed in period one ($\bar{x}_{1}$), is
\begin{align*}
\bar{x}_{2}= \frac{k\bar{x}_{1}+\sum_{j=1}^{n(\bar{x}_{1})}x_{2,j}}{k+n(\bar{x}_{1})},
\end{align*}
where $k\bar{x}_{1}$ is the sum of all observed payoffs in the first period. 
Thus,
\begin{align*}
E[\bar{x}_{2}|\bar{x}_{1}] = \frac{k\bar{x}_{1}+E[\sum_{j=1}^{n(\bar{x}_{1})}x_{2,j}|\bar{x}_{1}]}{k+n(\bar{x}_{1})}.
\end{align*}

Because $E[\sum_{j=1}^{n(\bar{x}_{1})}x_{2,j}|\bar{x}_{1}] = un(\bar{x}_{1})$, this can be written as
\begin{align*}
E[\bar{x}_{2}|\bar{x}_{1}] = \frac{k\bar{x}_{1}}{k+n(\bar{x}_{1})}+\frac{n(\bar{x}_{1})u}{k+n(\bar{x}_{1})}.
\end{align*}
Moreover, because $E(\bar{x}_{2}) = E_{\bar{x}_{1}}( E[\bar{x}_{2}|\bar{x}_{1}] )$ we get
\begin{align*}
E(\bar{x}_{2}) = E_{\bar{x}_{1}}(\frac{k\bar{x}_{1}}{k+n(\bar{x}_{1})})+E_{\bar{x}_{1}}(\frac{n(\bar{x}_{1})u}{k+n(\bar{x}_{1})} )
\end{align*}
Let $g(\bar{x}_{1}) = \frac{1}{k+n(\bar{x}_{1})}$. 
Because $E(\bar{x}_{1}g(\bar{x}_{1})) = E(\bar{x}_{1})E(g(\bar{x}_{1}))+cov(\bar{x}_{1},g(\bar{x}_{1}))$ and $E(\bar{x}_{1}) = ku$ we get
\begin{align*}
E_{\bar{x}_{1}}(\frac{k\bar{x}_{1}}{k+n(\bar{x}_{1})})= kuE(g(\bar{x}_{1}))+Cov(\bar{x}_{1},g(\bar{x}_{1})).
\end{align*}
Note that 
\begin{align*}
kE(g(\bar{x}_{1}))= E(\frac{k}{k+n(\bar{x}_{1})})= 1-E(\frac{n(\bar{x}_{1})}{k+n(\bar{x}_{1})} )
\end{align*}
Overall we get
\begin{align*}
E(\bar{x}_{2}) = u[1-E(\frac{n(\bar{x}_{1})}{k+n(\bar{x}_{1})}]\\
+Cov(\bar{x}_{1},g(\bar{x}_{1}))+uE(\frac{n(\bar{x}_{1}))}{k+n(\bar{x}_{1})} )\\
=  u+Cov(\bar{x}_{1},g(\bar{x}_{1})).
\end{align*}
$Cov(\bar{x}_{1},g(\bar{x}_{1}))$ is negative (positive) whenever $g(\bar{x}_{1})$ is a strictly decreasing (increasing) function of $\bar{x}_{1}$. Because $g(\bar{x}_{1}) = 1/(k+n(\bar{x}_{1}))$, which is a strictly decreasing function of $n(\bar{x}_{1})$, $Cov(\bar{x}_{1},g(\bar{x}_{1}))$ is negative whenever $n(\bar{x}_{1})$ is a strictly increasing function of $\bar{x}_{1}$ and $Cov(\bar{x}_{1},g(\bar{x}_{1}))$ is positive whenever $n(\bar{x}_{1})$ is a strictly decreasing function of $\bar{x}_{1}$.

\medskip

\noindent \textbf{Proof of Theorem 2:} 

Without loss generality we only consider cases when $u = 0$. A unimodal random variable symmetric around the mode $k$ can always be represented as $x = k+\varepsilon$ where $\varepsilon$ has a unimodal distribution symmetric around zero.  

We first prove the following Lemma

\medskip

\noindent \textbf{Lemma 1:} Let $x_{i}$, $i = 1,...,n$ be independent random variables with a unimodal distribution symmetric around the mode $0$. Suppose $r > 0$. Then $P(\sum_{i=1}^{k}x_{i} > r) > P(\sum_{i=1}^{v}x_{i} > r)$ whenever $k > v$. That is, a sum with more terms is more likely to be extreme than a sum with fewer terms. 

\medskip


\begin{figure}
\noindent \begin{centering}

\includegraphics[trim=2cm 12cm 4cm 2cm, clip=true, scale=0.65]{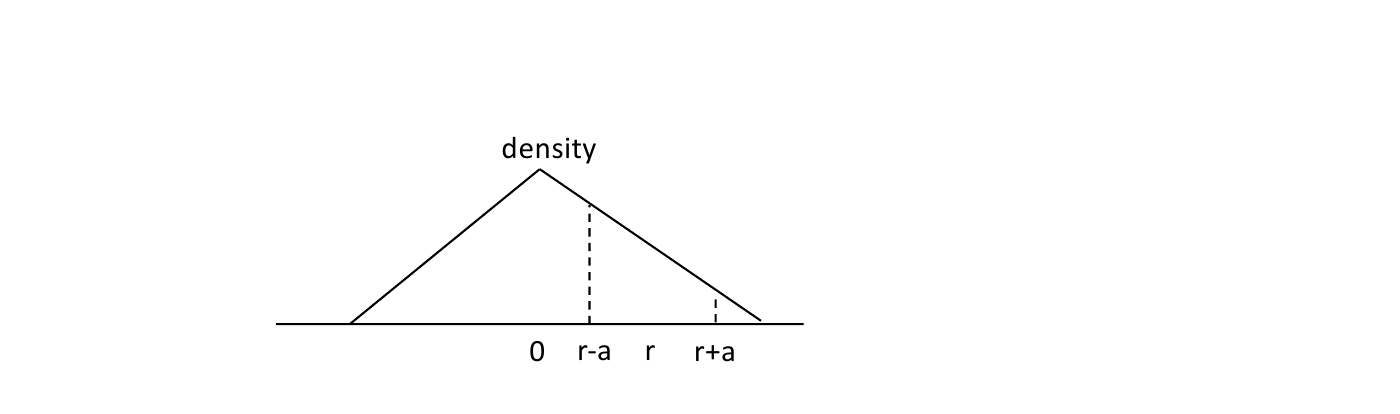}
\caption{Illustration of Proof}

\label{figeinhorn}

\par\end{centering}
\end{figure}


\noindent \textbf{Proof of Lemma 1:} Consider first the case when $k = 2$ and $v = 1$. Let $s_2 = x_1+x_2$. We need to show that $P(s_2 > r) > P(x_{1} > r)$, i.e., that adding $x_2$ to $x_1$ increases the chance that the sum is above $r$. Suppose first that $x_{1}= r+a$ where $a>0$. Then $s_2 > r$ if $x_{2} > r-x_1 = -a$. This occurs with probability $P(x_{2} > -a) =1-F(-a)$. Moreover, $1-F(-a) = F(a)$ due to symmetry around zero.  
Suppose next $x_{1}= r-a$ where $a>0$. Then $s_2 > r$ if $x_{2} > r-x_1 = a$. This occurs with probability $1-F(a)$. Altogether, $P(s_2 > r)$ equals
\begin{align*}
P(s_2 > r) = \int_{a=0}^{\infty}[f(r+a)F(a)+f(r-a)(1-F(a))]da.
\end{align*}
Now, $f(r+a)F(a)+f(r-a)(1-F(a)) > f(r+a)F(a)+f(r+a)(1-F(a))$ when $a > 0$ because $f(r-a) > f(r+a)$ when $a >0$, $r > u$, and $f(\cdot)$ has a unimodal distribution centered on zero. This is illustrated in Figure 1. Next, note that 
\begin{align*}
 f(r+a)F(a)+f(r+a)(1-F(a)) = f(r+a). 
\end{align*}
Moreover, $\int_{a=0}^{\infty}f(r+a)da = P(x_{1} > r)$. Thus, $P(s_2 > r) > P(x_1 > r)$. 

The proof now follows by induction, because the sum of $k$ independent random variables with a unimodal distribution symmetric around $u$ is also unimodal symmetric centered on $u$ (\cite{sorder}, page 173).

\medskip

Let now $r$ denote the the sum of the observed payoffs in the first period: $r = \sum_{j=1}^{k}x_{1,j}$. Because the sum of $k$ independent random variables with a unimodal distribution symmetric around $u$ is also unimodal symmetric centered on $u$ (Shaked and Shantikumar, 1996, p. 173), $r$ has a unimodal symmetric distribution centered on zero. 

Consider a fixed value of $r$. Suppose $r > 0$. The average after the second period becomes less than zero whenever 
\begin{align*}
\bar{x}_{2}= \frac{r+\sum_{j=1}^{n(r)}x_{2,j}}{k+n(r)} < 0,
\end{align*}
i.e., when $\sum_{j=1}^{n(r)}x_{2,j} < -r$. Here $n(r)$ is the sample size, for a given level of $r$. 
This occurs with probability $P(\sum_{j=1}^{n(r)}x_{2,j} < - r)$. Because the sum is also symmetrically distributed around zero, this equals $P(\sum_{j=1}^{n(r)}x_{2,j} > r)$.

Suppose next $r < 0$. The average after the second period remains below zero whenever 
\begin{align*}
\bar{x}_{2}= \frac{r+\sum_{j=1}^{n(r)}x_{2,j}}{k+n(r)} < 0,
\end{align*}
i.e., when $\sum_{j=1}^{n(r)}x_{2,j} < -r$. This occurs with probability $P(\sum_{j=1}^{n(r)}x_{2,j} < -r)$.

The probability that the average is below zero after the second period is thus
\begin{align*}
0.5\int_{r=0}^{\infty}f_{r}(r|r>0)P(\sum_{j=1}^{n(r)}x_{2,j} > r)dr+0.5\int_{r=-\infty}^{0}f_{r}(r|r<0)P(\sum_{j=1}^{n(r)}x_{2,j} < -r)dr.
\end{align*}
Now, $f_{r}(r|r>0) = f_{r}(r)/P(r>0) =  f_{r}(r)/0.5$. Similarly, $f_{r}(r|r<0) = f_{r}(r)/P(r<0) =  f_{r}(r)/0.5$. The probability that the average is below zero after the second period can thus be written as
\begin{align*}
\int_{r=0}^{\infty}f_{r}(r)P(\sum_{j=1}^{n(r)}x_{2,j} > r)dr+\int_{r=-\infty}^{0}f_{r}(r)P(\sum_{j=1}^{n(r)}x_{2,j} < -r)dr.
\end{align*}
By a change of variables $s = -r$, and noting that $f_{r}(-r) = f_{r}(r)$ for a distribution symmetric around zero, the last term can be written as
\begin{align*}
\int_{r=-\infty}^{0}f_{r}(r)P(\sum_{j=1}^{n(r)}x_{2,j} < -r)dr = \int_{s= 0}^{\infty}f_{r}(s)P(\sum_{j=1}^{n(-s)}x_{2,j} < s)ds 
\end{align*}
Altogether, the probability that the average is below zero after the second period equals
\begin{align*}
\int_{r=0}^{\infty}f_{r}(r)P(\sum_{j=1}^{n(r)}x_{2,j} > r)dr+\int_{r=0}^{\infty}f_{r}(r)P(\sum_{j=1}^{n(-r)}x_{2,j} < r)dr.
\end{align*}
Moreover, $P(\sum_{j=1}^{n(-r)}x_{2,j} < r) = 1-P(\sum_{j=1}^{n(-r)}x_{2,j} > r)$.
The probability that the average is below zero after the second period thus equals
\begin{align*}
\int_{r=0}^{\infty}f_{r}(r)dr + 
\int_{r=0}^{\infty}f_{r}(r)P(\sum_{j=1}^{n(r)}x_{2,j} > r)dr-\int_{r=0}^{\infty}f_{r}(r)P(\sum_{j=1}^{n(-r)}x_{2,j} > r)dr.
\end{align*}
Now, $P(\sum_{j=1}^{n(r)}x_{2,j} > r)$ is larger (smaller) than $P(\sum_{j=1}^{n(-r)}x_{2,j} > r)$ for all $r> 0$ when $n(x)$ is an increasing (decreasing) function of $x$. Thus, the probability that the average is below zero after the second period is larger (smaller) than
\begin{align*}
\int_{r=0}^{\infty}f_{r}(r)dr = 0.5.
\end{align*}

 \bigskip

\noindent \textbf{Proof of Theorem 3:} 
The belief after all payoffs observed in both periods one and two, $z_{2,n(z_{1,k})}$, given the belief at the end of period one ($z_{1,k}$), is equal to
\begin{align*}
z_{1,k}(1-b)^{n(z_{1,k})}+x_{2,1}b(1-b)^{n(z_{1,k})-1}+...+bx_{2,n(z_{1,k})}.
\end{align*}
The conditional expected value $E[z_{2,n(z_{1,k})}|z_{1,k}]$ equals
\begin{align*}
z_{1,k}(1-b)^{n(z_{1,k})}+E[x_{2,1}b(1-b)^{n(z_{1,k})-1}+...+bx_{2,n(z_{1,k})}|z_{1,k}].
\end{align*}
Because $E[x_{2,j}|z_{1,k}] = u$, this can be written as
\begin{align*}
E[z_{2,n(z_{1,k})}|z_{1,k}]=z_{1,k}(1-b)^{n(z_{1,k})}+(1-(1-b)^{n(z_{1,k})})u.
\end{align*}
Moreover, because $E(Y) = E_{X}( E[Y|X] )$, $E[z_{2,n(z_{1,k})}]$ equals
\begin{align*}
E_{z_{1,k}}[z_{1,k}(1-b)^{n(z_{1,k})}]+E_{z_{1,k}}[(1-(1-b)^{n(z_{1,k})})]u.
\end{align*}
Let $g(z_{1,k}) = (1-b)^{n(z_{1,k})}$. 
Because $E(z_{1,k}g(z_{1,k})) = E(z_{1,k})E(g(z_{1,k}))+Cov(z_{1,k},g(z_{1,k}))$, and $E(z_{1,k}) = u$, we get
\begin{align*}
E[z_{2,n(z_{1,k})}] = uE(g(z_{1,k}))+Cov(z_{1,k},g(z_{1,k})) + u(1-E(g(z_{1,k}))\\
=  u+Cov(z_{1,k},g(z_{1,k})).
\end{align*}
$Cov(z_{1,k},g(z_{1,k}))$ is negative (positive) whenever $g(z_{1,k})$ is a strictly decreasing (increasing) function of $\bar{x}_{1}$. Because $g(z_{1,k}) = (1-b)^{n(z_{1,k})}$, which is a strictly decreasing function of $n(z_{1,k})$, $Cov(z_{1,k},g(z_{1,k}))$ is negative whenever $n(z_{1,k})$ is a strictly increasing function of $z_{1,k}$ and $Cov(z_{1,k},g(z_{1,k}))$ is positive whenever $n(z_{1,k})$ is a strictly decreasing function of $z_{1,k}$.  

\medskip

\noindent \textbf{Proof of Theorem 5:} 
Without loss of generality we focus on the case when $m = 0$. If $m$ differs from zero, the distribution of payoffs is identical to a constant, $m$, plus a random variable with a normal distribution with mean zero.  
Let $S_1$ be the sum of the observed payoffs in period one. This sum is equally likely to be positive or negative. Suppose  $S_1 = z > 0$. It follows that the belief after the first period, $E[u|S_1 ] = (1/k)S_1\sigma^{2}_{u}/(\sigma^{2}_{u}+\sigma^{2}_{e}/k)$, is positive. Denote this belief by $b_1$, i.e., $b_1 = E[u|S_1 ] $. Note that $b_1$ is a strictly increasing function of $S_1$. 
This positive belief turns into a negative belief after period 2 whenever 
\begin{equation*}
b_{2} = \frac{\frac{1}{k+n(b_1)}(S_1+S_2)\sigma^{2}_{u}}{\sigma^{2}_{u}+\frac{\sigma^{2}_{e}}{k+n(b_1)}} < 0,
\end{equation*}
i.e., whenever $S_2 < -S_1$. We seek the probability of this event given the observed value of $S_1 = z$, i.e., we seek $Pr(S_2 < -S_1 | S_1 = z)$.

Conditional on $S_1 = z$, the sum of the payoffs in the second period, $S_2$, is a normally  distributed random variable with mean $n(z)E(u|S_1 = z) = n(z)z\sigma^{2}_{u}/(\sigma^{2}_{u}+\sigma^{2}_{e}/k)$ and variance $n(z)^{2}Var(u|S_1 = z) +\sigma^{2}_{e}n(z)$, where $Var(u_i|x_1 = z) = \sigma^{2}_{u}\sigma^{2}_{e}/(k\sigma^{2}_{u}+\sigma^{2}_{e})$ (see Lindgren (1993, p. 289). 
It follows that 
\begin{equation*}
Pr(S_2 < -S_1| S_1 = z) = \Phi(\frac{-z-n(z)(z/k)\sigma^{2}_{u}/(\sigma^{2}_{u}+\sigma^{2}_{e}/k)}{\sqrt{n(z)^{2}\sigma^{2}_{u}\sigma^{2}_{e}/(k\sigma^{2}_{u}+\sigma^{2}_{e})+\sigma^{2}_{e}n(z)}}).
\end{equation*}
Altogether, the probability that a positive belief after period one turns into a negative belief after period two equals
\begin{align*} 
Pr(+ \to -) = \\
\int_{z = 0}^{z = + \infty}\Phi(\frac{-z-n(z)(z/k)\sigma^{2}_{u}/(\sigma^{2}_{u}+\sigma^{2}_{e}/k)}{\sqrt{n(z)^{2}\sigma^{2}_{u}\sigma^{2}_{e}/(k\sigma^{2}_{u}+\sigma^{2}_{e})+\sigma^{2}_{e}n(z)}})f(z|z>0)dz.
\end{align*}
Because $f(z|z>0) = f(z)/P(z>0) = f(z)/0.5$, $Pr(+ \to -)$ equals
\begin{equation} 
 \int_{z = 0}^{z = + \infty}\Phi(\frac{-z-n(z)(z/k)\sigma^{2}_{u}/(\sigma^{2}_{u}+\sigma^{2}_{e}/k)}{\sqrt{n(z)^{2}\sigma^{2}_{u}\sigma^{2}_{e}/(k\sigma^{2}_{u}+\sigma^{2}_{e})+\sigma^{2}_{e}n(z)}})2f(z)dz,
\end{equation}
where $f(z)$ is the density of the sum of the payoffs in the first period which is a normal distribution with mean zero.

Suppose next $S_1 = z < 0$ implying that $E[u|S_1 ] < 0$. 
This negative belief turns into a positive belief after period 2
whenever $S_2 > -z$. 
Following the same reasoning as above $Pr(S_2 > -S_1| S_1 = z)$ equals 
\begin{equation*}
1-\Phi(\frac{-z-n(z)(z/k)\sigma^{2}_{u}/(\sigma^{2}_{u}+\sigma^{2}_{e}/k)}{\sqrt{n(z)^{2}\sigma^{2}_{u}\sigma^{2}_{e}/(k\sigma^{2}_{u}+\sigma^{2}_{e})+\sigma^{2}_{e}n(z)}}).
\end{equation*}
Because  $1-\Phi(y) = \Phi(-y)$, this can be written as
\begin{equation*}
\Phi(\frac{z+n(z)(z/k)\sigma^{2}_{u}/(\sigma^{2}_{u}+\sigma^{2}_{e}/k)}{\sqrt{n(z)^{2}\sigma^{2}_{u}\sigma^{2}_{e}/(k\sigma^{2}_{u}+\sigma^{2}_{e})+\sigma^{2}_{e}n(z)}}).
\end{equation*}
Altogether, the probability that a negative belief after period one turns into a positive belief after period two, $Pr(- \to +)$, equals
\begin{equation*} 
\int_{z = -\infty}^{z = 0}\Phi(\frac{z+n(z)(z/k)\sigma^{2}_{u}/(\sigma^{2}_{u}+\sigma^{2}_{e}/k)}{\sqrt{n(z)^{2}\sigma^{2}_{u}\sigma^{2}_{e}/(k\sigma^{2}_{u}+\sigma^{2}_{e})+\sigma^{2}_{e}n(z)}})2f(z)dz.
\end{equation*}
After the variable substitution, $z = -y$, this integral equals
\begin{equation*} 
\int_{y = 0}^{y = \infty}\Phi(\frac{-y-n(-y)(y/k)\sigma^{2}_{u}/(\sigma^{2}_{u}+\sigma^{2}_{e}/k)}{\sqrt{n(-y)^{2}\sigma^{2}_{u}\sigma^{2}_{e}/(k\sigma^{2}_{u}+\sigma^{2}_{e})+\sigma^{2}_{e}n(-y)}})2f(y)dy.
\end{equation*}
where we have used the fact that $f(-y) = f(y)$. 

We wish to show that $Pr(+ \to -) > Pr(- \to +)$, which requires us to show that 
\begin{align*}
\Phi(\frac{-z-n(z)(z/k)\sigma^{2}_{u}/(\sigma^{2}_{u}+\sigma^{2}_{e}/k)}{\sqrt{n(z)^{2}\sigma^{2}_{u}\sigma^{2}_{e}/(k\sigma^{2}_{u}+\sigma^{2}_{e})+\sigma^{2}_{e}n(z)}})   > \\
\Phi(\frac{-z-n(-z)(z/k)\sigma^{2}_{u}/(\sigma^{2}_{u}+\sigma^{2}_{e}/k)}{\sqrt{n(-z)^{2}\sigma^{2}_{u}\sigma^{2}_{e}/(k\sigma^{2}_{u}+\sigma^{2}_{e})+\sigma^{2}_{e}n(z)}}),
\end{align*}
over the positive domain of $z$. To do so, note first that the derivative of
\begin{equation*} 
g(z,w) = \frac{-z-wz\sigma^{2}_{u}/(\sigma^{2}_{u}+\sigma^{2}_{e}/k)}{\sqrt{w^{2}\sigma^{2}_{u}\sigma^{2}_{e}/(k\sigma^{2}_{u}+\sigma^{2}_{e})+\sigma^{2}_{e}w}},
\end{equation*}
with respect to $w$ is
\begin{equation*} 
\frac{\delta g(z,w)}{\delta w} = \frac{z}{2w\sqrt{\frac{\sigma^{2}_{e}w(\sigma^{2}_{e}+\sigma^{2}_{u}+\sigma^{2}_{e}w)}{\sigma^{2}_{u}+\sigma^{2}_{e}}}},
\end{equation*}
which is positive whenever $z > 0$. Hence $g(z,w)$ is an increasing function of $n(z)$. 
Because $\Phi(x)$ is a increasing function of $x$ and $n(z)$ is an increasing function of $z$, implying that $n_{2}(z) > n_{2}(-z)$, it follows that $\Phi(g(z,n(z)) > \Phi(g(z,n(-z))$ implying $Pr(+ \to -) > Pr(- \to +)$.

\bibliographystyle{apacite}



\bibliography{sampbib}



\end{document}